\documentclass[runningheads]{llncs}
\usepackage[T1]{fontenc}
\usepackage{graphicx}
\usepackage{orcidlink}

\begin{document}
\title{Leveraging OpenFlamingo for Multimodal Embedding Analysis of C2C Car Parts Data}
\titlerunning{OpenFlamingo for Multimodal Embeddings of Car Parts Data}
%
\author{Maisha Binte Rashid~\orcidlink{0009-0009-4781-5593} \and
Pablo Rivas~\orcidlink{0000-0002-8690-0987}}
\authorrunning{M.B. Rashid and P. Rivas}
%
\institute{Department of Computer Science, Baylor University, Texas, USA
\email{\{Maisha\_Rashid1,Pablo\_Rivas\}@Baylor.edu}}
\maketitle              
\begin{abstract}
In this paper, we aim to investigate the capabilities of multimodal machine learning models, particularly the OpenFlamingo model, in processing a large-scale dataset of consumer-to-consumer (C2C) online posts related to car parts. We have collected data from two platforms, OfferUp and Craigslist, resulting in a dataset of over 1.2 million posts with their corresponding images. The OpenFlamingo model was used to extract embeddings for the text and image of each post. We used $k$-means clustering on the joint embeddings to identify underlying patterns and commonalities among the posts. We have found that most clusters contain a pattern, but some clusters showed no internal patterns. The results provide insight into the fact that OpenFlamingo can be used for finding patterns in large datasets but needs some modification in the architecture according to the dataset. 

\keywords{Multimodal machine learning \and secure and trustworthy cyberspace \and multimodal embeddings.}
\end{abstract}
\section{Introduction}\label{sec1}

The combination of text and images on online platforms provides a rich information tapestry that pushes the limits of classical natural language processing and machine learning. Large-scale datasets with text and images are becoming more widely available, making it possible to create effective multimodal machine-learning models. These models use the complementary information found in both textual and visual data to perform a range of tasks, such as visual question answering~\cite{manmadhan2020visual} and image captioning~\cite{sharma2020image}. Our exploration of this intricate area prompted us to investigate multimodal machine learning models, concentrating on OpenFlamingo~\cite{awadalla2023openflamingo}, which is an open-source replica of the Flamingo model~\cite{alayrac2022flamingo}.

This work uses OpenFlamingo to analyze a new dataset of consumer-to-consumer online posts related to automotive parts. With its combination of images and related text, this dataset offers a rich multimodal signal that complex machine-learning algorithms can utilize. Our objective is to determine whether these multimodal models can process and differentiate between various posts in the dataset, which encompass a broad spectrum of automotive components and related parts. Using OpenFlamingo, we extract embeddings from each post's image and text components. Next, we use clustering algorithms for these joint embeddings to determine whether similar posts can be grouped. This gives us insight into how well the model can represent the underlying semantic relationships between the multimodal data. By analyzing the clustering's performance on this actual dataset, we hope to shed insight into the capabilities and constraints of multimodal machine learning techniques in handling large-scale, heterogeneous datasets.

By conducting this study on a large-scale dataset, we aim to evaluate the robustness and scalability of multimodal models such as OpenFlamingo in processing heterogeneous, real-world data. The knowledge gathered from this study can help create more successful multimodal machine learning methods that can be applied to a variety of fields.

\section{OpenFlamingo}\label{sec2}
OpenFlamingo is an open-source framework with a variety of models from 3B to 9B parameters that makes it easier to train large autoregressive vision-language models~\cite{awadalla2023openflamingo}. The models use a frozen vision encoder (CLIP ViT-L/14) and pre-trained, frozen language models. The term "frozen" indicates that the parameters of these models are not updated during training. Instead, they remain fixed as they were initialized from their pretrained states. A cross-attention module predicts the next token while simultaneously cross-attending to the vision encoder's outputs. Using publicly available components, OpenFlamingo's stack trains on web-scraped image-text sequences from open-source datasets like Multimodal C4 and LAION-2B, which is a big step towards democratizing research on autoregressive vision-language models. 
In the initial pre-processing stage, the authors designate the positions of images within the textual sequence using \texttt{<image>} tokens. Additionally, the authors append tokens following the textual content after an image. For instance, in a sequence where \texttt{x} represents an image, denoted as \texttt{x This is a cat}, it would be transformed into \texttt{<image> This is a cat}.

\section{Methodology}\label{sec3}
\begin{figure}[h]
\centering
  \includegraphics[width=\textwidth]{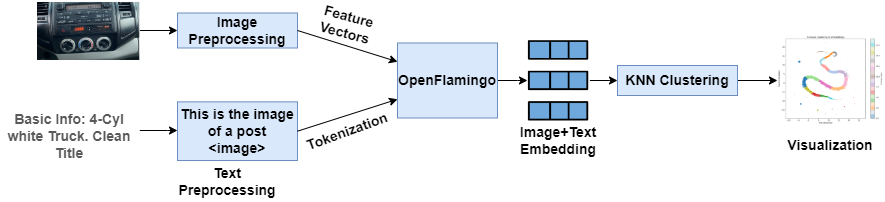}
  \caption{A diagram of our methodology. Note that the input is multimodal: text and images.}
  \label{fig:diagram}
\end{figure}

In order to conduct our research, we assembled a large dataset from OfferUp and Craigslist, two well-known consumer-to-consumer websites. After carefully searching these sites for content related to car parts, a significant amount of data was gathered for analysis. Fig. \ref{fig:diagram} shows the process of how we conducted this research. In the later sections, we discuss each step in detail. 
\subsection{Data Collection}\label{subsec2}
\begin{figure}[t]
  \includegraphics[width=\columnwidth]{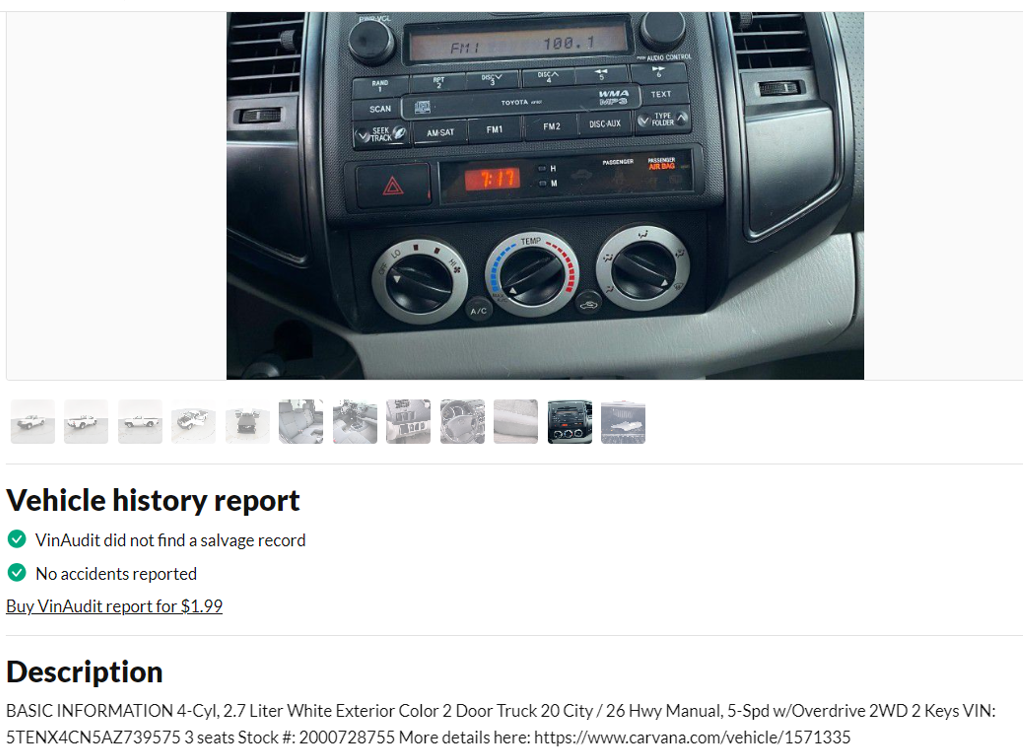}
  \caption{Example of a post related to car parts from OfferUp. Used with permission.}
  \label{fig:offerup}
\end{figure}

The data acquisition process was carried out through web scraping techniques on publicly available data, targeting posts related to car parts on both Craigslist and OfferUp. From OfferUp, we extracted a total of 650,654 posts along with their corresponding images, amounting to approximately 500GB of data. Similarly, we gathered 637,679 posts and their associated images from Craigslist, which cumulatively approximated 50GB. In both these websites each post has multiple images associated with it. This extensive collection formed the basis of our dataset for subsequent analysis. An example of a post from OfferUp is shown in Fig.~\ref{fig:offerup}.

\subsection{Data Preprocessing and Tokenization}
In the subsequent preprocessing phase, an initial step involved the tokenization of the posts. We introduced specific phrases to enhance the semantic structure of the data. For every post, the phrase "\texttt{This is a post}" was prefixed, serving as a standard identifier for the beginning of a post. For posts lacking associated images, the statement "\texttt{no image added with this post}" was appended at the end to denote the absence of visual content. Conversely, for posts accompanied by a single image, the phrase "\texttt{This is the image that goes with the post}" was used. In cases where multiple images were associated with a post, we utilized the phrase "\texttt{These are the images that go with the post}," followed by the insertion of the \texttt{<image>} token corresponding to the number of images attached to the post. This structured approach to tokenization was instrumental in preparing the data for the embedding phase.
\subsection{Embedding Extraction}
Using the OpenFlamingo model, we obtained embeddings for each post's text and image components. Using these embeddings, we can represent each post as a single vector in a high-dimensional space, capturing the semantic and visual information found in the multimodal data. 
\subsection{Clustering}  
After retrieving the embeddings from OfferUp and Craigslist posts, we set out to categorize them to identify any patterns or shared features. For categorization, we utilized the $k$-means algorithm, widely recognized for its effectiveness in distinguishing distinct groups within large datasets~\cite{macqueen1967some}. Our aim was to establish 20 clusters, facilitating a detailed analysis that may inform on specific characteristics of each group of posts. Because our dataset is huge and comprises a variety of auto parts, we have decided to select cluster number 20. 

Following the categorization, our next objective was to simplify the visualization of these groupings. To achieve this, we employed UMAP, a technique designed for reducing data complexity while preserving its essential structure, making it more manageable to visualize~\cite{mcinnes2018umap}. UMAP's strength lies in its ability to reduce high-dimensional data into a more comprehensible form.
\begin{figure}[h!]
  \includegraphics[width=\textwidth]{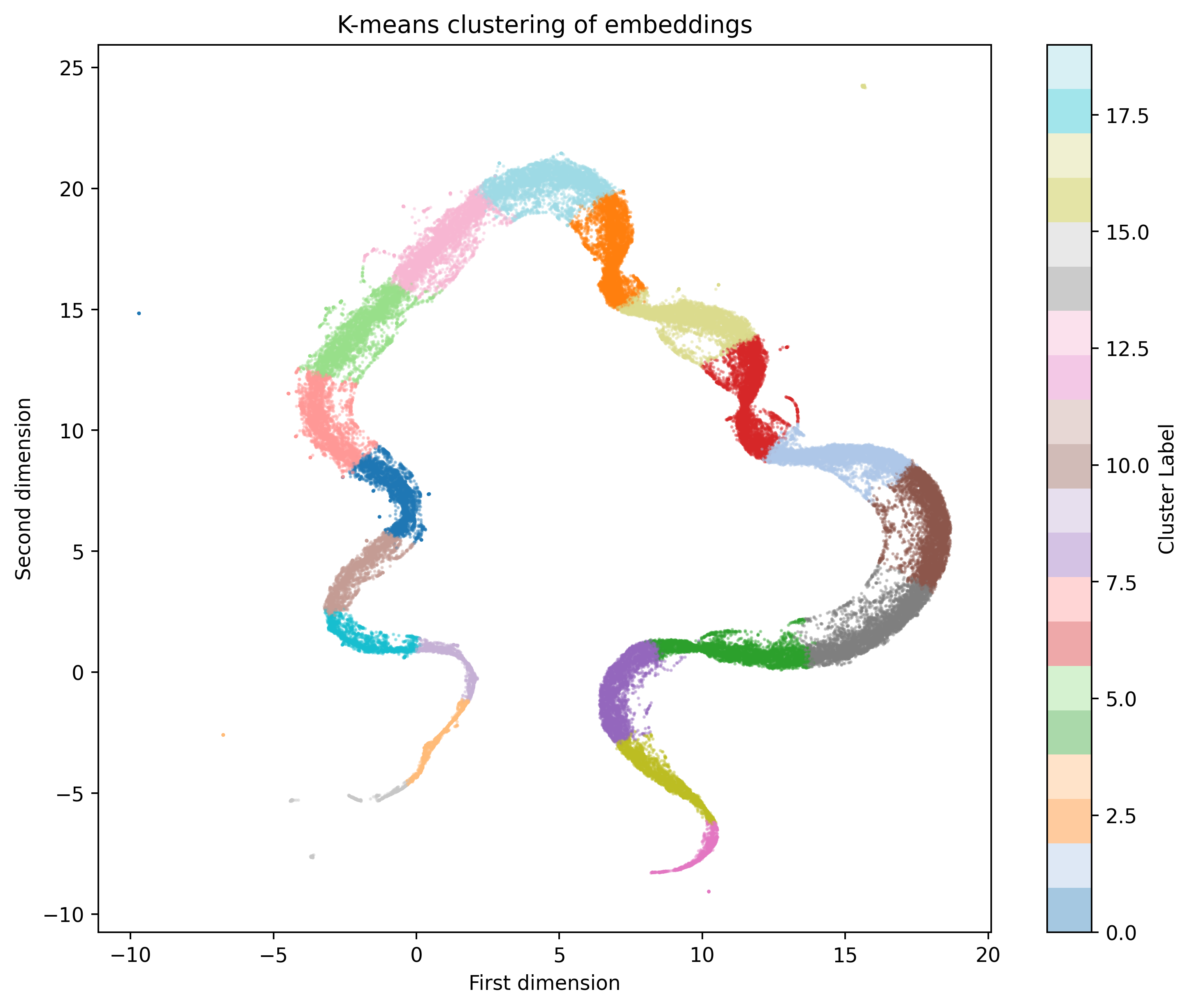} \caption {$k$-means clustering visualization on CraigsList samples.}
  \label{fig:kmeansa}
\end{figure}
\begin{figure}[h!]
  \includegraphics[width=\textwidth]{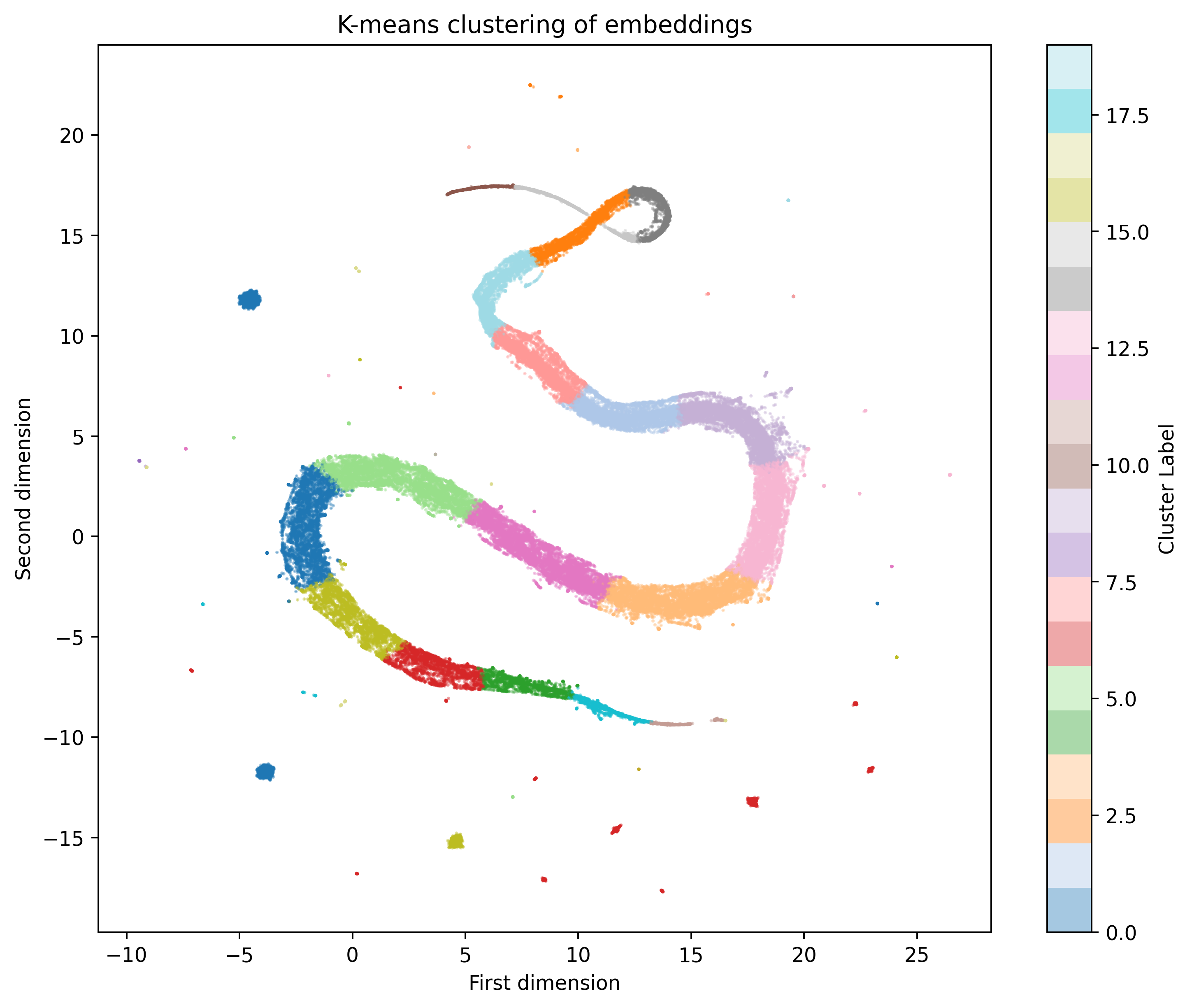}
  \caption {$k$-means clustering visualization on OfferUp samples.}
  \label{fig:kmeansb}
\end{figure}
This sequential process—first categorizing the posts into clusters using the $k$-means algorithm, then applying UMAP for visualization—ensured that our analysis remained rigorous and our results accessible. It allowed us to work with the data in its original complexity for categorization, subsequently simplifying it for presentation purposes, thereby making our findings understandable and engaging for a broader audience.

For the visualization stage, we chose 70,000 samples at random from each of our datasets. This sample size was considered reasonable for visualization yet large enough to accurately reflect the diversity of the entire dataset. After that, we plotted the dimensionality-reduced embeddings, using each point to represent a post and its assigned cluster among the 20. This gave us a visual representation of the cluster distribution and separation. Fig.~\ref{fig:kmeansa} presents a scatter plot visualization of the embeddings derived from Craigslist data, and Fig.~\ref{fig:kmeansb} corresponds to the OfferUp data. This graphical representation effectively illustrates the clustering of posts based on their embeddings, which we discuss next. 
\section{Result and Discussion}\label{sec4}

To analyze the clustering of posts from OfferUp and Craigslist, we looked at the top 10 posts closest to the centroids of each cluster. By examining the 10 posts that are closest to each centroid, we can more accurately infer the behaviors and characteristics associated with each centroid. This approach ensures that our analysis captures the most significant patterns within the dataset. This investigation aimed to identify any observable patterns or features shared by the posts categorized under each cluster.
In the Craigslist dataset, the clustering revealed distinct categories of car parts:
\begin{itemize}
    \item \textbf{First Cluster:} Outer body parts and car seats
    \item \textbf{Second Cluster:} Tires and trucks
    \item \textbf{Third Cluster:} Inner car parts 
    \item \textbf{Forth Cluster:} Posts with similar endings, often including phone numbers or phrases like 'If you need other parts, call me.'
\end{itemize}
Other clusters focused on similar car parts. However, three clusters did not show clear patterns, indicating some randomness in post categorization.

The OfferUp dataset exhibited similar characteristics:
\begin{itemize}
    \item \textbf{First Cluster:} Tires
    \item \textbf{Second Cluster:} Head and tail lights
    \item \textbf{Third Cluster:}  Colorful body parts 
    \item \textbf{Forth Cluster:} Seats and covers
\end{itemize} 
While both platforms showed some common trends, there was considerable variation, with some posts not aligning with the typical characteristics of their clusters.

One important finding from our investigation is the possible influence of the OpenFlamingo model's architecture, specifically its focus on processing one image for each post. This feature may make it more difficult for the algorithm to classify posts with several photos, which may explain some of the observed cluster randomness and unpredictability. By improving the model's capacity to interpret and categorize posts with multiple visual elements, we could achieve more coherent and consistent clusters.

\section{Conclusion}\label{sec5}

The study's findings provide insight into the OpenFlamingo model's strengths and weaknesses for handling large, diverse datasets of consumer-to-consumer online posts. After performing $k$-means clustering on the image and text embedding space, the results indicate that
most clusters showed clear patterns associated with particular automotive parts or qualities. However, some of the data ended up in clusters with no obvious pattern. This result indicates that although the OpenFlamingo model shows promise in identifying trends in large, diverse datasets, there is still space for improvement, especially in handling posts with several photos.

By leveraging the OpenFlamingo model to analyze multimodal data from consumer-to-consumer car parts posts, the study demonstrates the potential of advanced machine learning techniques to handle large-scale, heterogeneous datasets. This research can be beneficial to provide a framework for their application in various fields such as e-commerce, automotive, and online marketplaces. In summary, this work contributes to the broader goal of creating robust and scalable machine learning systems capable of extracting valuable insights from complex, real-world data.

\begin{credits}
\subsubsection{\ackname} Part of this work was funded by the National Science Foundation under grants CNS-2210091, CHE-1905043, and CNS-2136961.

\subsubsection{\discintname}
The authors have no competing interests to declare that are
relevant to the content of this article.
\end{credits}
%
%
%
\bibliographystyle{splncs04}
\bibliography{biblio}

\end{document}